\documentclass[letterpaper]{article} 
\usepackage{aaai23}  
\usepackage{times}  
\usepackage{helvet}  
\usepackage{courier}  
\usepackage[hyphens]{url}  
\usepackage{graphicx} 

\usepackage{booktabs}
\usepackage{algorithm}
\usepackage{algorithmic}
\usepackage{amsmath}
\usepackage{natbib}  
\usepackage{caption} 
\usepackage{balance}
\usepackage{algorithm}
\usepackage{algorithmic}

\usepackage{newfloat}
\usepackage{listings}
\DeclareCaptionStyle{ruled}{labelfont=normalfont,labelsep=colon,strut=off} 
\floatstyle{ruled}
\newfloat{listing}{tb}{lst}{}
\floatname{listing}{Listing}
\title{Uncertainty-Aware Image Captioning}
\author{
    Zhengcong Fei, Mingyuan Fan, Li Zhu, Junshi Huang\thanks{The corresponding author.}\\ Xiaoming Wei, Xiaolin Wei
}
\affiliations{
    Meituan\\

    Beijing, China\\
    \{name\}@meituan.com
}

\usepackage{bibentry}

\begin{document}

\maketitle

\begin{abstract}
It is well believed that the higher uncertainty in a word of the caption, the more inter-correlated context information is required to determine it. However, current image captioning methods usually consider the generation of all words in a sentence sequentially and equally. In this paper, we propose an uncertainty-aware image captioning framework, which parallelly and iteratively operates insertion of discontinuous candidate words between existing words from easy to difficult until converged. We hypothesize that high-uncertainty words in a sentence need more prior information to make a correct decision and should be produced at a later stage. The resulting non-autoregressive hierarchy makes the caption generation explainable and intuitive. Specifically, we utilize an image-conditioned bag-of-word model to measure the word uncertainty and apply a dynamic programming algorithm to construct the training pairs. During inference, we devise an uncertainty-adaptive parallel beam search technique that yields an empirically logarithmic time complexity. Extensive experiments on the MS COCO benchmark reveal that our approach outperforms the strong baseline and related methods on both captioning quality as well as decoding speed. 
\end{abstract}

\section{Introduction}


Image captioning, which aims to generate textual descriptions of input images, is a critical task in multimedia analysis \cite{stefanini2021show}.
Previous works in this area are mostly based on an encoder-decoder paradigm \cite{Vinyals2015Show,Xu2015Show,Rennie2017Self,Anderson2017Bottom,huang2019attention,cornia2020meshed,pan2020x,fei2022attention,li2022comprehending,yang2022reformer}, where a convolution-neural-network-based image encoder first process an input image into visual representations, and then a recursive-neural-network or Transformer-based language decoder produces a corresponding caption based on these extracted features. The generation process usually relies on a chain-rule factorization and is performed in an autoregressive manner, \emph{i.e.}, words by words from left to right. Although this paradigm brings performance superiority, high latency during inference becomes a grave disadvantage in some real-time applications. To this end, non-autoregressive image captioning (NAIC) models are proposed to improve decoding speed by predicting every word in parallel \cite{nat,fei2019fast,fei2021partially}. This advantage comes at a sacrifice on performance since modeling next word is trickier when not conditioned on sufficient contexts.

\begin{figure}
	\begin{center}
		\includegraphics[width=1\columnwidth]{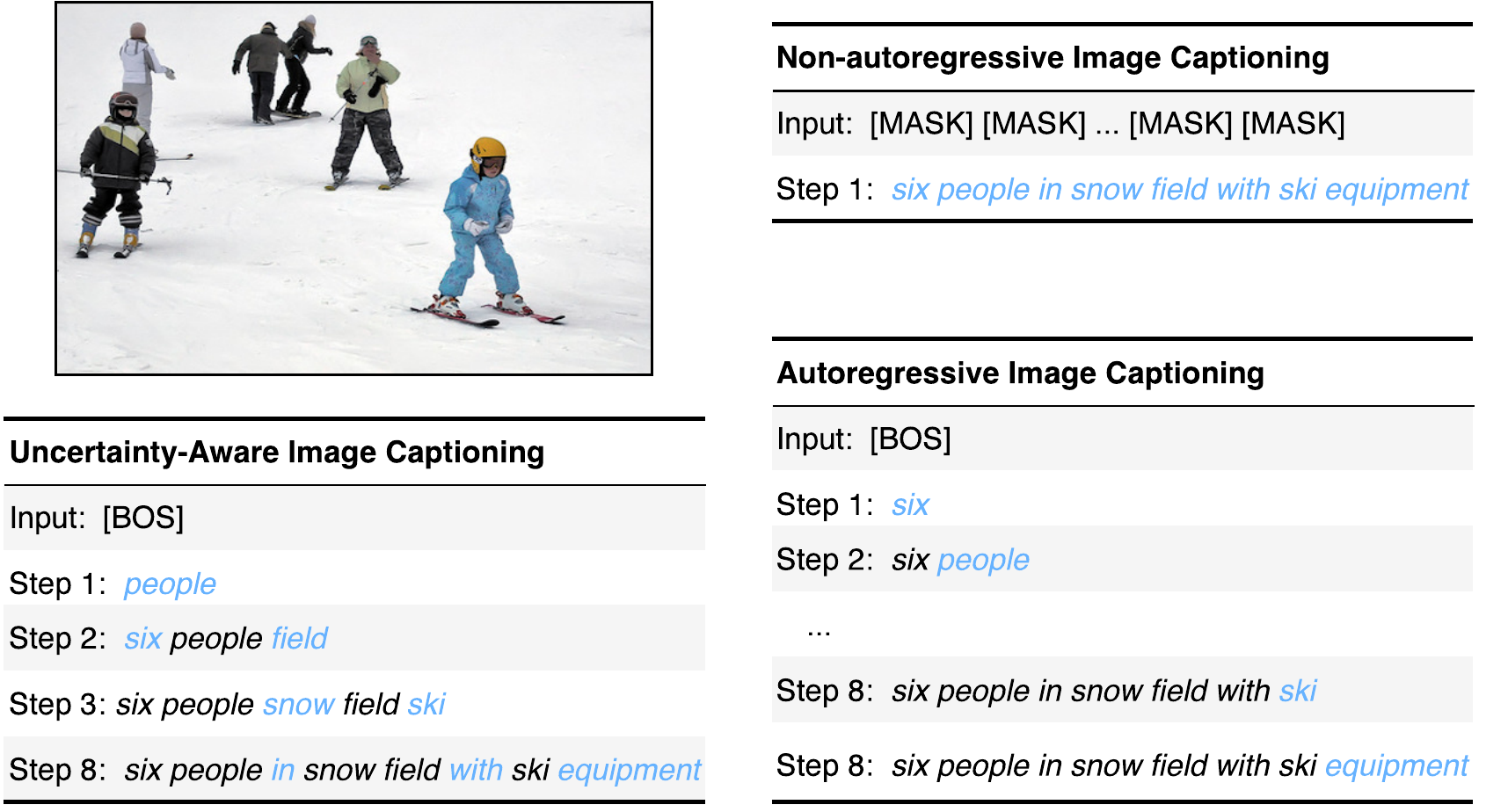}
	\end{center}
	\caption{Illustration of autoregressive, non-autoregressive, and uncertainty-aware image captioning. AIC model generates the next word conditioned on the given image and all preceding subsentence, while NAIC model outputs all words in one shot. Comparatively, UAIC considers a caption generation in several stages and various discontinuous words are generated parallelly in each stage. words in blue indicate newly generated words at the current stage. Interestingly, UAIC allows informative words (\emph{e.g., people, field}) generated before the no-informative words (\emph{e.g., in, with}). }
	\label{fig:1}
\end{figure}

Intensive efforts have been devoted to non-autoregressive image captioning to seek a better trade-off between inference time and captioning performance. Generally, approaches can be roughly divided into two lines. The first line of work leverages the iterative decoding framework to break the independence assumption, which first generates an initial caption and then refines iteratively by taking both the given image and the results of the last iteration as input \cite{gao2019masked,dir}. The other line of work tries to modify the Transfomer structure to better capture dependency and position information by leveraging extra autoregressive layers in the decoder \cite{fei2019fast,guo2020non}. Besides, \cite{fei2020iterative} introduces latent variables to eliminate the modal gap and develop a more powerful probabilistic framework to simulate more complicated distributions. \cite{yan2021semi} splits the captions into word groups averagely and produces the group synchronously. 
In addition to parallel generation, a range of semi-autoregressive models \cite{wang2018semi,ghazvininejad2020semi,stern2019insertion,gu2019levenshtein,fei2021partially,fei2022deecap,fei2022efficient,zhou2021semi} pay attention to non-monotonic sequence generation with limited forms of autoregressiveness, \emph{i.e.}, tree-like traversal, which are mainly based on the insertion operation. However, all these image captioning methods treat all words in a sentence equally and ignore the generation completeness between them, which does not match the actual human-created situation.

In this paper, we propose an \textbf{U}ncertainty-\textbf{A}ware parallel model for faster and better \textbf{I}mage \textbf{C}aptioning, referred to as UAIC. As illustrated in Figure \ref{fig:1},  given an image, the UAIC model first produces low-uncertainty keywords, \emph{i.e.}, objective from the image, and then inserts details of finer granularity from easy to difficult in stages. This process iterates until a caption is finally completed. 
Note that the generation of discontinuous words in each stage is synchronous and parallel. 
To measure the word uncertainty in a sample, we introduce a bag-of-words (BoW) model \cite{zhang2010understanding} according to image-only conditional distribution. The intuition behind this is that the higher the cross-entropy and uncertainty have in a word, the harder it is to generate, and more word dependency information is required to determine it \cite{heo2018uncertainty,kaplan2018uncertainty,zhou2020uncertainty}. Meantime, a dynamic programming-based algorithm is applied to split instances into ordered data pairs for the training process. Under such a scenario, we can say that the training sample is in an uncertainty-aware manner and the proposed model can implicitly learn the effective word generation order. We also integrate an uncertainty-adaptive beam search following uncertainty change for decoding acceleration. We believe that our proposed approach is simple to understand and implement, yet powerful, and can be leveraged as a building block for future works.
The main \emph{contirbutions} of this paper are summarized as follows: 
\begin{itemize}
\item  We propose a new uncertainty-aware model for parallel image caption generation. Compared with previous work, our model allows difficulty control on the generation and enjoys a significant reduction of empirical time complexity from $\mathcal{O}(n)$ to $\mathcal{O}(\text{log } n)$ at best. 
\item We introduce an uncertainty estimation model inspired by the idea of bag-of-word. Based on the word-level uncertainty measurement, a heuristic dynamic programming algorithm is applied to construct the training set. 
\item We devise an uncertainty-adaptive beam search customized to our approach, which dynamically adjusts the beam size to trade-off accuracy and efficiency. 
\item Experiments on the MS COCO benchmark demonstrate the superiority of our UAIC model over strong baselines. In particular, to improve reproducibility and foster new research in the field, we publicly release the source code and trained models of all experiments.
\end{itemize}

\section{Background}

\paragraph{Image caption generation.} Given an image $I$, image captioning aims to generate sentence $S= \{w_1, \ldots, w_T\}$ to describe the visual content. Specifically, the autoregressive image captioning (AIC) model factorizes the joint distribution of $S$ in a standard left-to-right manner, \emph{i.e.}, 
\begin{equation}
	p(S|I)=\prod_{t=1}^Tp(w_t|S_{<t}, I).
\end{equation}
While AIC models have achieved great success in terms of caption quality, the time consumption during inference is still far away from satisfactory \cite{nat,gao2019masked,guo2021non}. 
Correspondingly, a line of works begins to develop into one-shot non-autoregressive image captioning (NAIC) models. These models break the autoregressive dependency by decomposing the joint probability with:
\begin{equation}
	p(S|I) = \prod_{t=1}^T p(w_t| I).
\end{equation}
However, the loss of autoregressive dependency largely hurts the consistency of the output sentences, increases the difficulty in the learning process, and thus leads to a low-quality translation.

\paragraph{Insertion transformer.} A new sequence generation via insertion operation is introduced from  \cite{stern2019insertion,quach2020blank}. Generally, in each step, a bi-direction insertion transformer is performed on the expanded sequence to compose the representation for each candidate slot in between every two consecutive positions. After that, an optimization process for the joint distribution of position-word is performed to support an insertion-based text generation. 
Practically, the insertion transformer is a modification of the original transformer decoder architecture. The original decoder predicts the next word based on the previously generated sequence while the insertion transformer can predict words for all the available slots. In this setup, words on the decoder side can attend to the entire sequence instead of only their left side. This means the causal self-attention mask is removed in the original decoder. 
Formally, on step $t$ where a word $w_{i}$ is inserted in between position $i$ and $i+1$ of historical context $C_t = \{I, \tilde{S}\}$, where $\tilde{S}$ denotes the partially generated sentence, the log step likelihood can be written as:
\begin{equation}
\begin{split}
    \text{log }p(w_{i}, \phi_{i}| C_t)  = &\text{ log } p_{pos}( \phi_{i} | C_t) \\
    & + \text{ log } p_{word}(w_{i}|C_t, \phi_i ),
\end{split}
\end{equation}
where $p_{pos}$ and $p_{word}$ stand for the insertion position prediction and the $i$-th word distribution of bi-directional sequence encoding. The expectation of the negative log step likelihood overall permitted context-insertion pairs at each step is computed as the step loss. The step losses from the first step to the last one are summed up as the final sequence loss.

\section{Methodology}

\begin{figure*}
	\begin{center}
		\includegraphics[width=1.8\columnwidth]{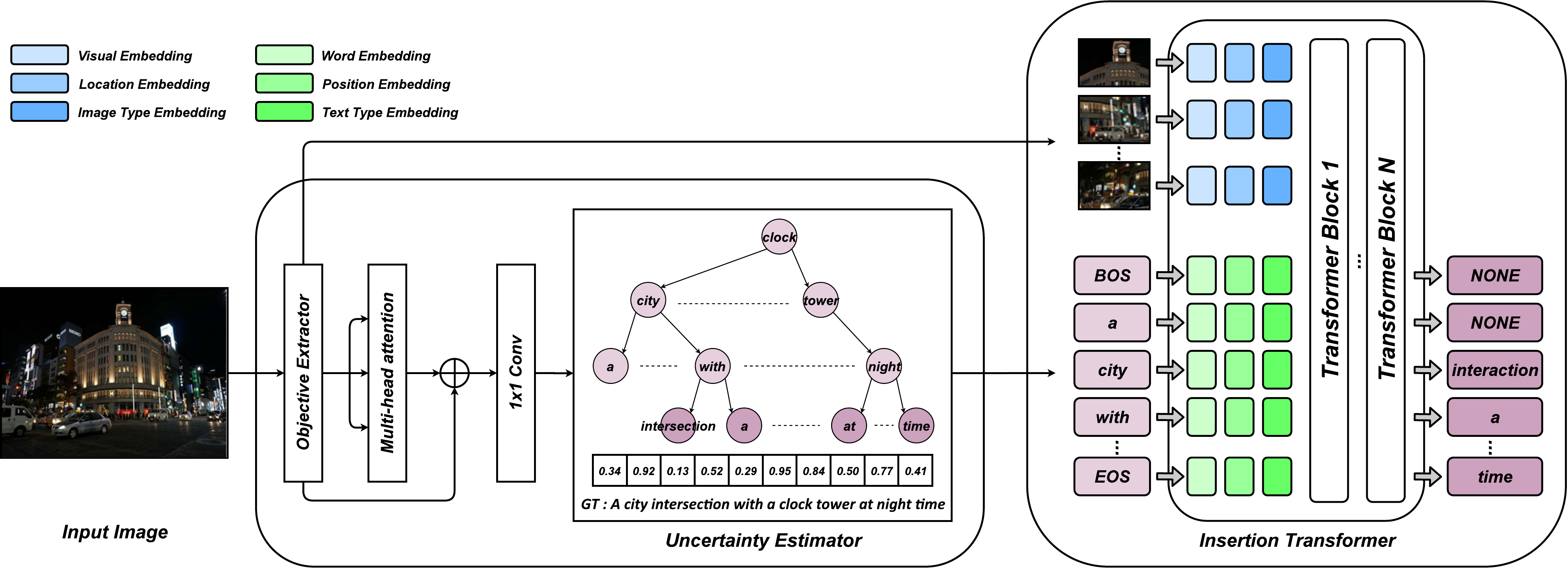}
	\end{center}
	\caption{
	Overview of UAIC framework. Image regions and caption words are projected into the same dimensional space by the sum of three embeddings, respectively. All inputs are then combined together through $N$-layer Transformer blocks. The final hidden states $h$ are fed into a word classifier to predict the new insertion words or no operation word [NONE]. The training data preparation reverses the above generation process through the uncertainty estimator, which takes images as input and outputs the uncertainty scores of each word to determine the generation order of caption words.
	}
	\label{fig:2}
\end{figure*}

\subsection{Overview}

The image captioning process of the UAIC model can be formulated as a series of $K$ stages:
$\{S_1, \ldots, S_K\}$, such that for each $k \in \{2, \ldots, K\}$, $S_{k-1}$ is a subsequence of  $S_k$, \emph{i.e.}, the following stage can be perceived as a finer-resolution sequence compared to the sequence of preceding stage. $S_K$ is the final generation, under the condition that iterative refinement is converged.
At each step, at most one new word can be produced and inserted between two existing words. Formally, we can factorize the distribution according to the measurement uncertainty of each word as:
\begin{align} 
	p(S|I) = \prod_{k=1}^{K}p(S_k|S_{k-1}, I), \\
	p(S_k|S_{k-1}, I) = \prod_{w \in S_k -S_{k-1}} p(w|S_{k-1},I), \label{eq:4}
\end{align}
where $S_k -S_{k-1}$ is the word difference set between two sentences. 
Intuitively, under this condition, the easier and lower-uncertainty words, such as objective and attributes, appear in earlier stages, and the higher-uncertainty auxiliary words, such as prepositions, are generated at the later large. 
As illustrated in Figure \ref{fig:2}, our UAIC model is based on Insertion Transformer \cite{stern2019insertion,cornia2020meshed} and the training objective comprises both ($i$) likelihood of an insertion indicator for each slot (between two existing words), and ($ii$) the likelihood of each new word conditioning on the activated slot. 
To handle this case, we expand the vocabulary with a special no-operation symbol [NONE]. During inference time, the model can predict either a word from the vocabulary to insert, or a [NONE] symbol indicating no new word will be inserted at a certain slot at the current stage. By utilizing this special word, the two objectives are unified.

The \emph{\textbf{key difference}} for the AIC in model structure is that the UAIC removes the causal attention mask from the decoder, allowing fully contextualized output representations to be derived after each insertion. Meanwhile, compared with the iterative refinement-based non-autoregressive image captioning \cite{fei2020iterative}, UAIC model produces the sentence in several stages and does not need to predict the caption length before decoding, which can avoid length prediction errors effectively. 

\subsection{Uncertainty Estimator}

Uncertainty comes with judgment. For example, when we consider a word to be generated, we also know how confident we are in the judgment \cite{hama2019exploring}. 
The caption is composed of words, and words have the ability to express the visual content. For this reason, we construct a vocabulary $\mathcal{V}$ in which the words can sufficiently describe different images, and then, we utilize an image-conditioned bag of words (BoW) \cite{wu2010semantics} based on this vocabulary that is directly related to the ground-truth captions,
\begin{equation}
	\pi = -p(\mathcal{V} |I) = -[p(w_i|I; w_i \in \mathcal{V})].
\end{equation}
Specifically, given a set of visual features extracted from input image $I$, we utilize a multi-head self-attention Transformer module \cite{attention} to fully exploit the semantic information. This module calculates attention vectors for these regional features, then the convex combinations of these features under different attentions are concatenated for the final representation construction.
In the last step, an MLP layer takes the hidden states as input and generates an uncertainty score $\pi \in[-1,0]$ for each word $w_i$ in vocabulary $\mathcal{D}$. The vector of these scores is concatenated to formulate the final uncertainty evaluation, which embeds the information of the objects, their attributes, and relationships into the image representation.
For the training of uncertainty measurement model, the target generated by mapping function $\psi$ is a binary vector and each element denotes whether a word exists in the caption valued 1 or not valued 0. To simplify, the words existing in the caption should have a lower uncertainty score. Here we use the MSE loss to update the model parameters of the uncertainty measure as: 
\begin{equation}
	L_{UE} = ||\pi + \psi(S)||^2.
\end{equation}

Please note that collecting all the existing words is unnecessary as the redundant words do not relate to image content. To cut down the dictionary scale, we introduce a stop word list, i.e., only include the descriptive words, which can be categorized into two groups. Here we select the words for objects from the image captioning dataset as the first group and the words for describing object attributes and relationships as the second group. In this way, we can determine the final size of the word dictionary $\mathcal{V}$.

\subsection{DP-based Data Pair Construction}

After computing the word-level uncertainty for each instance, we can construct pairs of image-conditioned text sequences at adjacent stages, \emph{i.e.}, $(S_{k-1}, I; S_k)$, as the model input and labels, for training convenience. As a result, each sample $(S, I)$ from training dataset is broken into a consecutive series of pairs $(S_{1}, I; S_2), \ldots, (S_{K-1}, I; S_K)$. Two properties are desired when constructing the training instance: ($i$) low-uncertainty words should be generated in an earlier stage to receive more dependency information. ($ii$) in the later stage, multiple discontinuous words can be generated parallel to accelerate the inference process.

Here we leverage the Insertion Transformer, which allows at most one new word to be generated between every two existing words, and sentence length at most doubles at each iteration. Consequently, the optimal number of iterations $K$ is log($T$), where $T$ is the caption length. However, forcing generation of a balancing tree will ignore the uncertainty of each word. To this end, we incorporate an approach to mask the caption by considering both word uncertainty and efficiency using dynamic programming (DP). To accommodate the nature of insertion-based generation, the masking procedure is under the constraint that no consecutive words can be masked at the same stage. Under such a condition, we score each word and select a subset of words that add up to the lowest uncertainty. This allows the algorithm to adaptively choose as many low-uncertainty words as possible to mask. 
Formally, as an integer linear programming problem \cite{vanderbei2015linear,chan2020imputer}, the objective is to find an optimal masking pattern $\Phi = \{ \phi_1, \ldots, \phi_T \}$ to minimize the uncertainty score, where $\phi_t \in \{0, 1\}$, and $\phi_t = 1$ represents discarding the corresponding word $w_t$ with uncertainty $u_t$, and $\phi_t = 0$ indicates remaining. For a sentence $S$, the objective can be formulated as:
\begin{equation}
    \begin{split}
        \text{min} \sum_{t=1}^T \phi_t \cdot u_t, \label{eq:7} \\
    \text{s.t.   } \phi_t \phi_{t+1} \neq 1, \forall t
    \end{split}
\end{equation}
Since Equation \ref{eq:7} is computationally expensive, we employ dynamic programming shown in Algorithm 1 to optimize the computation referred to as \cite{takaoka2002efficient}.



\begin{algorithm}[tb] 
\small
	\label{a:1}
	\caption{DP-based Training Data Pair Construction }  
	\textbf{Input:} Estimated uncertainty list $\pi$ \\
	\textbf{Output:} Masking pattern $\Phi$
	\begin{algorithmic}[1]
	 \STATE $\triangleright$ Initialization\\ 
	  \STATE $V \leftarrow \{v_0=0, v_1=u_1, v_2=0, ..., v_L=0\}$\\
	 \STATE $\Phi \leftarrow \{\phi_0=\{\}, \phi_1=\{1\}\}$\\
	  \STATE $t \leftarrow 2$\\
	 \WHILE{t $\leq$ T}
	    \IF{$v_{t-1} > v_{t-2}+u_t$}
	   \STATE     $v_{t}=v_{t-1}$\\
	  \STATE      $\phi_{t}=\phi_{t-1}$\\
	   \STATE     $\phi_{t}$ \textbf{append} $0$\\
	    \ELSE
	    \STATE    $v_{t}=v_{t-2}+u_t$\\
	   \STATE     $\phi_{t}=\phi_{t-2}$\\
	  \STATE      $\phi_{t}$ \textbf{append} $1$\\
	    \ENDIF
	    \ENDWHILE
    \STATE \textbf{return} $\Phi [1:]$
	 \end{algorithmic}
\end{algorithm} 


Another thing to care for is once in a stage $S_K$, all the slots predict [NONE] for the next stage, the generation procedure is converged and $S_K$ is the final output sequence. To account for this final stage $S_K$, during data preparation we incorporate an $(S_K, I; \text{[NONE]}*L)$ pair for each sentence in the training data, where the target sentence denotes a sequence of [NONE] with the same length of $S_K$. To enable the model to insert at the beginning and the end of the sequence, an [BOS]
word and a [EOS] word are added at the beginning and at the end of each sentence, respectively.

\begin{table*}[t]
	\begin{center}
	\setlength{\tabcolsep}{1.5mm}{
		\begin{tabular}{l|cccccc|cc}
			\hline
			Models&BLEU-1&BLEU-4&METEOR&ROUGE&CIDEr&SPICE&Latency&SpeedUp\\
			\hline \hline
			\multicolumn{9}{l}{\emph{Autoregressive image captioning}} \\
			\hline
			NIC-v2 \cite{Vinyals2015Show} &-&32.1&25.7&-&99.8&-&-&-\\
			Up-Down \cite{Anderson2017Bottom}  &79.8&36.3&27.7&56.9&120.1&21.4&-&-\\
			AoANet$^\dagger$ \cite{huang2019attention} &\text{80.2}&\text{38.9}&\text{29.2}&\text{58.8}&\text{129.8}&\text{22.4}&-&-\\ 
			M2-T$^\dagger$\cite{cornia2020meshed} &80.8&39.1&29.2&58.6&131.2&22.6&-&-\\
			RSTNet$\dagger$ \cite{zhang2021rstnet}&81.1&39.3&29.4&58.8&133.3&23.0&-&-\\
			DIFNet$\dagger$ \cite{wu2022difnet}&81.7&40.0&29.7&59.4&136.2&23.2&-&-\\
			\hline
			\multicolumn{9}{l}{\emph{Non-autoregressive image captioning}} \\  \hline
			MNIC$^{\dagger \P}$ \cite{gao2019masked} &75.4&30.9&27.5&55.6&108.1&{21.0}&-&2.80$\times$\\ 
			FNIC$^\dagger$ \cite{fei2019fast} &-  &36.2&27.1 & 55.3&115.7&20.2&{\text{-}}&{\text{8.15}}$\times$\\ 
			MIR$^{\dagger \P}$  \cite{dir} &- &32.5&27.2&55.4&109.5&20.6&-&1.56$\times$\\ 
			CMAL$^\dagger$ \cite{guo2020non} &80.3 &37.3&28.1&58.0&124.0&21.8&-&13.90$\times$\\ 
			IBM$^{\dagger \P}$ \cite{fei2020iterative} &77.2 &36.6&27.8&56.2&113.2&20.9&-&3.06$\times$\\
			SAIC$\dagger$ \cite{yan2021semi}
			&80.3&38.4&29.0&58.1&127.1&21.9&-&3.42$\times$\\
			\hline
			\multicolumn{9}{l}{\emph{Uncertainty-aware image captioning}} \\  \hline
			UAIC$^\dagger$ &80.9&38.8&29.2&58.7&131.7&22.8&55ms&3.18$\times$\\
			\hline
		\end{tabular} }
	\end{center}
	{\caption{Performance comparisons of different captioning models using different evaluation metrics on the MS COCO karpathy test set. All values except Latency and SpeedUp are reported as a percentage (\%). “$\dagger$” denotes the model is based on Transformer architecture. AIC is our implementation of autoregressive teacher model, which has the same structure as UAIC. The SpeedUp values of NAIC models are from the corresponding paper. $\P$ represents the NAIC model is iterative refinement-based.}
		\label{tab:1}}
\end{table*}

\subsection{Training Objective}

With all the data pairs $( S_{k-1}, I;  S_k)$ created as described above to serve as the model input and output, we can optimize the model with the following objective:
\begin{equation}
\small
	\begin{split}
		\mathcal{L} &= -\text{log }p(S_{k}|S_{k-1},I), \\
		&= -\sum_{w \in S^+} \text{log } p(w| \phi_{k-1},S_{k-1},I)p(\phi_{k-1}|S_{k-1},I),
	\end{split}
\end{equation}
where word difference set $S^+ = S_{k}-S_{k-1}$, and $\phi_{k-1}$ denotes an indicator function for slots in the $k$-th stage, representing whether an insertion operator should be applied in a slot.
The same insertion transformer module is re-used at different stages. We empirically observe that the UAIC model can learn to insert different words at different stages under corresponding condition. 

\subsection{Uncertainty-Adaptive Beam Search}

Generally, UAIC model generates captions stage-by-stage from easy to difficult using greedy or beam search, by applying Insertion Transformer repeatedly until no additional word is generated. If a [NONE] word is generated, it is deleted at the next round.
According to Equation \ref{eq:4}, all new words are simultaneously generated based on the existing words at the previous stage. 
Inspired by the fact that when predicting easy and certain words at an early stage, we can adopt a smaller beam size for caption generation; otherwise, when producing more complete words, a larger beam size is expected.
To this end, we introduce an uncertainty-adaptive beam search algorithm for decoding. Specifically, our method applies an approximate local beam search at each iteration to find the optimal stage-wise decoding. At the $t$-th slot of $k$ stage, our method first generates top $B_k$ word candidates by applying one evaluation step based on existing generation. Prediction is limited to these $B_k$ word candidates, and thus beam search procedure
as described above is applied on the narrow band of $B_k$ instead of the full vocabulary $\mathcal{V}$. The uncertainty-adaptive beam size can be set: 
\begin{equation}
	B_k = 3 + \text{int}(4*\text{max}(-0.5, \text{min}(0.5, \frac{u_{avg}-u_k}{u_{avg}}) )),
\end{equation}
where $B_k$ and $u_k$ denote the beam size and subsentence uncertainty at stage $k$, $u_{avg}$ is the average uncertainty of the total training dataset.

\begin{table}
	\begin{center}
		\begin{tabular}{llc}
			\hline
			Model &Complexity&Acceleration\\
			\hline \hline
			AIC&$N(D+\Upsilon)+E$&1 \\
			NAIC&$D+\Upsilon+E$&$\approx N$\\
			IR-NAIC&$K(D+\Upsilon)+E$& $\approx \frac{N}{K}$ \\
			UAIC&$\text{log}N (D+\Upsilon)+E$& $\approx\frac{N}{\text{log }N}$\\
			\hline
		\end{tabular}
	\end{center}
			{\caption{Complexity and acceleration of different caption decoding approaches.}
			\label{tab:2}}
\end{table}

\subsection{Computation Complexity Analysis} 
Here, we also provide a theoretical analysis of time complexity and acceleration of different image caption decoding properties. 
As presented in Table \ref{tab:2}, $D$ denotes the time consumed on the decoder network, \emph{i.e.}, calculating a distribution over the target vocabulary at each time step and $\Upsilon$ denotes the time consumed on searching for top scores,  $E$ is the time consumed on the image feature extracting and encoding, $N$ denotes the average length of caption, and $K$ is the iterative refinement number. In practice, ($\textbf{i}$)  $D$ is usually much larger than $\Upsilon$ since the network is deep, and ($\textbf{ii}$) the proportion of $E$ in the total time procedure is pretty small since image feature encoding can be highly parallelized. As mentioned above, we can find that the speed bottleneck of image captioning model lies in its autoregressive decoding process and our proposed UAIC model can obtain an obvious caption decoding acceleration.

\begin{table*}[t]
\small
	\begin{center}
	\setlength{\tabcolsep}{2.5mm}{
		\begin{tabular}{lcccccccccccccc}
			\hline
			&\multicolumn{2}{c}{BLEU-1}&\multicolumn{2}{c}{BLEU-2}&\multicolumn{2}{c}{BLEU-3}&\multicolumn{2}{c}{BLEU-4}&\multicolumn{2}{c}{METEOR}&\multicolumn{2}{c}{ROUGE-L}&\multicolumn{2}{c}{CIDEr}\\ \hline
			&c5&c40&c5&c40&c5&c40&c5&c40&c5&c40&c5&c40&c5&c40\\
			\hline
			\hline
			Up-Down$^*$ &80.2&95.2&64.1&88.8&49.1&79.4&36.9&68.5&27.6&36.7&57.1&72.4&117.9&120.5\\
			M2-T$^*$  &81.6&96.0&66.4&90.8&51.8&82.7&39.7&72.8&29.4&39.0&59.2&74.8&129.3&132.1\\ 
			CMAL$^*$ &79.8&94.3&63.8&87.2&48.8&77.2&36.8&66.1&27.9&36.4&57.6&72.0&119.3&121.2\\ 
			SAIC &80.0 & 94.5&64.1&88.2&49.2&78.8&37.2&67.8&28.0&36.8&57.7&72.4&121.4&123.7\\ \hline
			UAIC$^*$ &81.9&96.3&66.5&91.1&51.8&83.0&39.6&72.9&29.2&38.9&59.2&74.7&129.0&132.8\\ 
			\hline
		\end{tabular}}
	{\caption{Leaderboard of various captioning models on the online MS COCO test server. $^*$ denotes the ensemble model.}
		\label{tab:3}}
	\end{center}
\end{table*}

\section{Experiments}


\subsection{Experimental Preparation}

\paragraph{Dataset.}
We evaluate our proposed method on MS COCO \cite{chen2015microsoft}, which is a standard benchmark for image captioning tasks.  To be consistent with previous works, \cite{huang2019attention,cornia2020meshed}, we adopted the Karpathy split \cite{karpathy2015deep} that contains 113,287 training images equipped with five human-annotated sentences each and 5,000 images for validation and test splits, respectively.  We also replicate the same data processing as previous works for fair comparisons, \emph{i.e.}, all the sentences are converted to lower cases, and we omit words that occur less than five times. 

\paragraph{Evaluation metrics.}
We utilize five standard automatic metrics simultaneously, namely  BLEU-$n$ \cite{Papineni2002BLEU}, METEOR \cite{Lavie2007METEOR}, ROUGE-L \cite{Flick2004ROUGE}, CIDEr \cite{dis1}, and SPICE \cite{spice}. 

\paragraph{Implementaion details.}
The proposed UAIC model closely follows the same network architecture and hyper-parameters setting as the conventional Transformer model as \cite{cornia2020meshed}. Specifically, the number of stacked blocks is 3, the hidden size is 512, and feed-forward filter size is 2048 with a 0.2 dropout rate. For visual representation, we follow the operation in \cite{wu2022difnet} using both the pre-extracted grid \cite{jiang2020defense} and segment features \cite{xiong2019upsnet} and employ an additional 3-layer multi-head self-attention layer for fusion. 
During training, we train the UAIC model for 15 epochs with an initial learning rate of 3e-5 and decay it by 0.9 every five epochs with the combined loss presented in Equation 9 \cite{he2019bag}. Adam \cite{kingma2014adam} optimizer with 3000 steps warm-up trick is employed.  The decoding time for speed-up estimation is measured on a single image without mini-batching and feature extraction, averaged over the whole test split with a 32G V100 GPU.

\subsection{Overall Performance}

\paragraph{Offline evaluation.}
We first compare the performance of our UAIC model against other non-autoregressive models and state-of-the-art autoregressive image captioning models. Among the autoregressive, AoANet, M2-T, and AIC are based on similar Transformer architecture as ours, while others are based on LSTM. 
Table \ref{tab:1} summarizes the performance of different image captioning models on the offline MS COCO dataset. According to the evaluation results, we can find that our uncertainty-aware model achieves significant improvements over the previous non-autoregressive image captioning models across most metrics, strikingly narrowing the performance gap between AIC. 
More encouragingly, not only NAIC models, but also we are supervised to find that UAIC can even surpass the AIC models trained from scratch in some metrics. We attribute it to two reasons: ($\textbf{i}$) UAIC is fine-tuned on a well-trained AIC model, making the training process easier and smoother, and ($\textbf{ii}$) generating caption in an increasing uncertainty order has a better guided future information than just training sequential from left to right.  Comparing speedups, our method obtains a significant speedup of more than a factor of 3 over the autoregressive counterpart, with latency reduced to about 55 ms. 
Remarkably, our UAIC is consistently achieving good quality with various beam search sizes. All these results demonstrate that our UAIC model is more promising compared with conventional AIC and NAIC models.

\paragraph{Online evaluation.}
We also submit an ensemble version of three UAIC models trained dependently to the official online testing server. 
The performance leaderboard on an official testing dataset with 5 reference captions (c5) and 40 reference captions (c40) is shown in Table \ref{tab:3}. 
According to the listed results, the proposed UAIC method maintains a compared performance compared with state-of-the-art autoregressive captioning models and obtains a substantial improvement in descriptive quality compared with strong non-autoregressive image captioning models.

\subsection{Ablation Study}
\paragraph{Effect of uncertainty-aware order.}

To evaluate the effectiveness of our uncertainty-aware generation order, we compare it with three other generation order training baselines under the same network architectures. As shown in Table \ref{tab:4}, we notice the performance drops when replacing the uncertainty-aware with the random or conventional sequential (from left to right), indicating that predicting words from easy to completeness in a sentence benefits the captioning quality. When training the model with anti-uncertainty order, i.e., training examples to the model in a complex-to-easy manner, we further observe a significant performance decrease, affirming the effectiveness of easy-to-complex uncertainty-aware generation manner again.

\begin{table}
	\begin{center}
		\setlength{\tabcolsep}{1.5mm}{
			\begin{tabular}{l|ccccc}
				\hline
				Order &B-4&M&R&C&S\\
				\hline
				\hline
				Sequential&38.5&29.0&58.5&129.2&22.6\\
				Random&38.6&29.0&58.6&129.6 (\textbf{0.4}$\uparrow$)&22.7\\ 
				Anti-Uncer.&38.4&28.6&58.2&127.5 (\textbf{1.7}$\downarrow$)&22.1\\
				Uncertainty&38.8&29.2&58.7&131.7 (\textbf{2.5}$\uparrow$)&22.8\\
				\hline
			\end{tabular}
			{\caption{Effect of different caption generation order on MS COCO test set. The results show the necessity of modeling the uncertainty-aware order.}
				\label{tab:4}}
		}
			\end{center}
\end{table}

\paragraph{Effect of dynamic beam search.}
As discussed in Section 3.5, dynamic beam search strategy is proposed to adjust beam size adaptively conditioned on the uncertainty of previously generated words, such that it can accelerate the decoding speed while avoiding a significant performance loss. Here, we investigate the effect of this strategy, and the results are presented in Table \ref{tab:5}. We can discover that: ($\textbf{i}$) as beam size grows larger, the performance of captioning, \emph{i.e.} CIDEr and BLEU-4 metrics, increases significantly, which is in line with the previous report; ($\textbf{ii}$)  the model equipped with the proposed dynamic beam search achieves a noteworthy acceleration, indicating the effectiveness of the beam size change and ($\textbf{iii}$)  through utilizing an uncertainty-adjust beam search strategy, we achieve a good balance between decoding speed and caption quality.

\begin{table}
	\begin{center}\setlength{\tabcolsep}{2mm}{
			\begin{tabular}{l|cccccr}
				\hline
				Beam size&B-4&M&R&C&S&Latency\\
				\hline
				\hline
				1&38.2&28.5&57.7&128.5&22.0&30ms\\
				3&38.6&28.8&58.2&129.8&22.4&52ms\\
				5&38.9&29.3&58.7&132.1&22.8&77ms\\
				Dynamic&38.8&29.2&58.7&131.7&22.8&55ms\\
				\hline
		\end{tabular}}
			\end{center}
		{\caption{Effect of uncertainty-adaptive beam search. Results are evaluated on MS COCO test set and demonstrates the effectiveness and efficiency.}
			\label{tab:5}}
\end{table}

\subsection{Case Study}
\begin{figure}[t]
	\centering
	\includegraphics[width=1.\columnwidth]{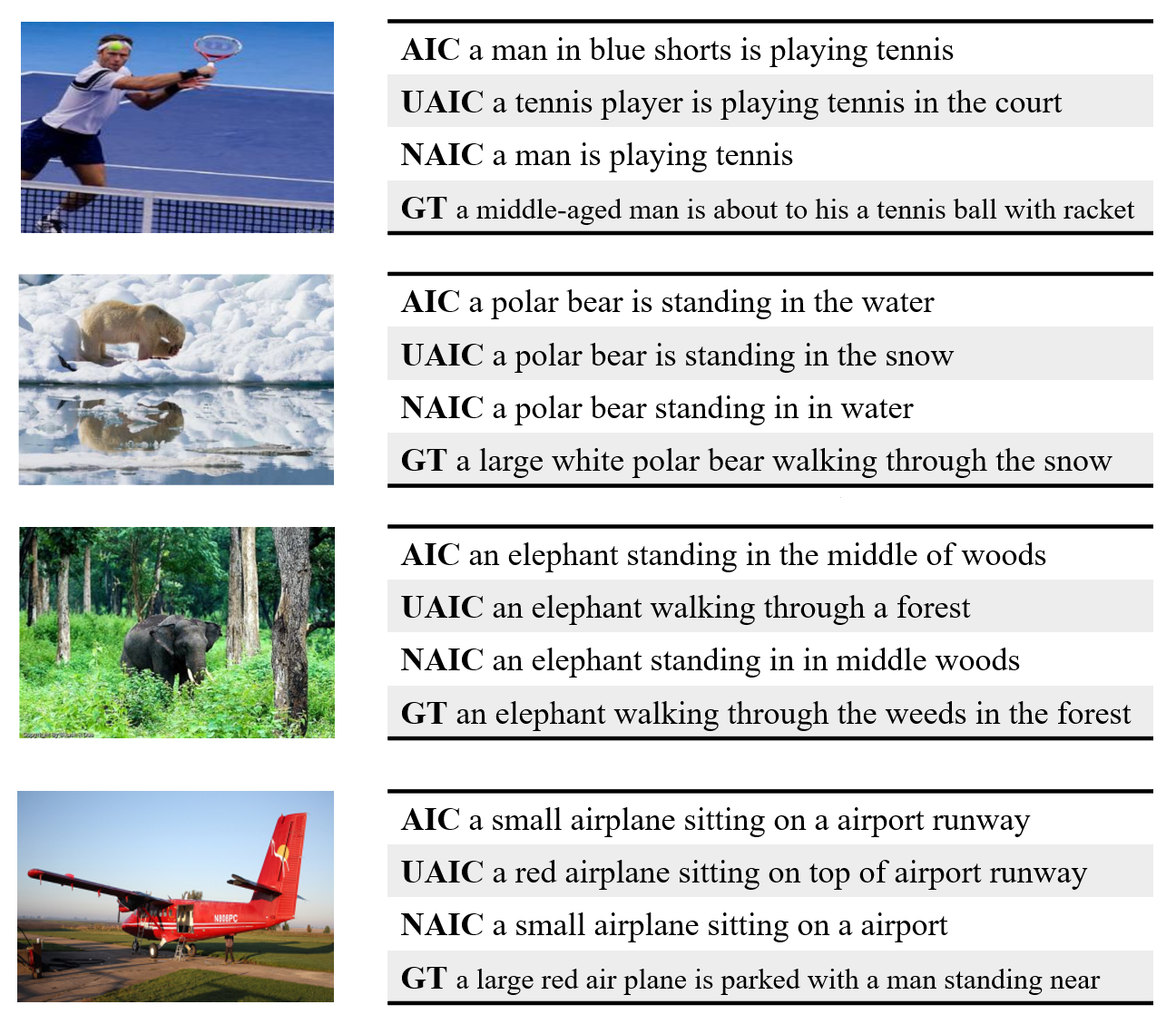}
	\caption{Examples of the generated captions from AIC, UAIC, and NAIC models with the same architectures. GT represents a human-annotated ground-truth caption. }
	\label{fig:4}
\end{figure}

To obtain a more intuitive perception, we present several examples of generated image captions from AIC, NAIC, and our proposed UAIC, which hold the same model architectures, coupled with human-annotated ground truth sentences (GT) in Figure \ref{fig:4}. As can be seen,  in general, all models hold the capability to reflect the content of the given image accurately. Meantime, the incoherent problem, including repeated words and incomplete content, is severe in the sentence generated by pure NAIC or AIC, while it can be effectively alleviated by UAIC, \emph{i.e.},  “snow” terms in the second sample. This again confirms that our proposed uncertainty-aware caption generation paradigm can leverage the captioning model to reduce word prediction errors.

\subsection{Human Evaluation}

To better evaluate the qualities of the generated captions by using different decoding approaches \cite{huang2019attention,exploring}, we also conducted a human study to compare our uncertainty-aware image captioning against two patterns, \emph{i.e.}, autoregressive and non-autoregressive image captioning with 15 workers.  In practice, we show one ground-truth sentence paired with a corresponding image generated by different approaches and asked: can you determine whether the given sentence has been generated by a system or by a human being? For each pairwise comparison, 200 images are randomly extracted from the Karpathy split of the testing set for them to compare. The results from annotators that pass the Turing test are listed as: the percentages of AIC, UAIC, and NAIC are 72.5\%, 76.8\%, and 60.5\% respectively. It demonstrates that the proposed UAIC model holds the superiority to provide accurate and human-like descriptions with well-designed generation order.

\section{Related Work}
With the advance in deep learning techniques,  encoder-decoder framework based on autoregressive decomposition has become the benchmark for image captioning task \cite{Vinyals2015Show,Xu2015Show,Anderson2017Bottom,Rennie2017Self,huang2019attention,cornia2020meshed,fei2021memory}. Recently, many attempts \cite{nat,wei2019imitation} have been made to generate words in a non-autoregressive manner to accelerate the inference. In other words, the caption generation process can be parallel or non-monotonic.
Specifically,  \emph{et al.} \cite{fei2019fast} first introduce non-autoregressive into image captioning, which reorders important words with a light RNN before the later parallel decoding. \cite{dir} refine with a post-processing masked step to remedy the conditional independence in the caption generation. \cite{guo2020non} addresses the inconsistency problem with a multi-agent learning paradigm and propose a sentence-level optimization objective and \cite{yan2021semi} proposes a fixed word group for parallel decoding. On the other hand, Insertion transformer proposed by \cite{stern2019insertion,gu2019levenshtein,zhang2020pointer} is a partially autoregressive generation model that can predict a sequence from middle to two sides, also known as balanced binary tree orders, providing fast inference while maintaining a good performance. 
Our method is built upon this approach and generalizes to explore a cross-modal application condition with learnable order.
In addition, all these models ignore the generation complexity and uncertainty of different words in a caption, which is not in line with reality. In contrast, our UAIC model incorporates it into the caption generation order and maintains an uncertainty-aware process.

\section{Conclusion}
In this paper, we propose a novel uncertainty-aware framework for image captioning.
Specifically, our method leverages an Insertion Transformer-based structure to generate a caption from easy to difficult in a non-autoregressive manner. 
An image-conditioned bag-of-words model is introduced to estimate the word-level uncertainty for training data, which is more explainable and comprehensive. 
A dynamic programming-based algorithm is then applied to construct the data pair based on the estimated uncertainty of each word. 
Additionally, we develop an uncertainty-adaptive beam search technique to further improve the decoding efficiency.  
Experimental results on the MS COCO benchmark show that our UAIC model can achieve comparable performance with state-of-the-art autoregressive counterparts, while at the same time enjoying $\sim$3 inference speedup. 
Moreover, our work opens a wide range of possibilities for a generation paradigm where monotonic ordering is not the most natural choice, and we would be excited to explore some other areas in future work.

\balance
\bibliography{cite}

\end{document}